\documentclass[10pt,twocolumn,letterpaper]{article}

\usepackage{cvpr}              

\usepackage{graphicx}
\usepackage{amsmath}
\usepackage{amssymb}
\usepackage{booktabs}

\usepackage{caption}
\usepackage[nolist]{acronym}
\begin{acronym}[PHOENIX14\textbf{T} ]

\acrodefplural{rnn}[RNNs]{Recurrent Neural Networks}
\acrodefplural{cnn}[CNNs]{Convolutional Neural Networks}
\acrodefplural{hmm}[HMMs]{Hidden Markov Models}
\acrodefplural{gru}[GRUs]{Gated Recurrent Units}
\acrodefplural{crf}[CRFs]{Conditional Random Fields}
\acrodefplural{gan}[GANs]{Generative Adversarial Networks}
\acrodefplural{gat}[GATs]{Graph Attention Networks}
\acrodefplural{gcn}[GCNs]{Graph Convolutional Networks}
\acrodefplural{gpu}[GPUs]{Graphic Processing Units}
\acrodefplural{gnn}[GNNs]{Graph Neural Networks}

\acrodefplural{mdn}[MDNs]{Mixture Density Networks}

\acrodefplural{vae}[VAEs]{Variational Autoencoders}

\acro{adain}[AdaIN]{Adaptive Instance Normalisation}

\acro{bsl}[BSL]{British Sign Language}
\acro{bleu}[BLEU]{Bilingual Evaluation Understudy}
\acro{blstm}[BLSTM]{Bidirectional Long Short-Term Memory}
\acro{cnn}[CNN]{Convolutional Neural Network}
\acro{crf}[CRF]{Conditional Random Field}
\acro{cslr}[CSLR]{Continuous Sign Language Recognition}
\acro{ctc}[CTC]{Connectionist Temporal Classification}
\acro{dl}[DL]{Deep Learning}
\acro{dgs}[DGS]{German Sign Language - Deutsche Gebärdensprache}
\acro{dsgs}[DSGS]{Swiss German Sign Language - Deutschschweizer Geb\"ardensprache}
\acro{dtw}[DTW]{Dynamic Time Warping}
\acro{fc}[FC]{Fully Connected}
\acro{ff}[FF]{Feed Forward}
\acro{gat}[GAT]{Graph Attention Network}
\acro{gan}[GAN]{Generative Adversarial Network}
\acro{gcn}[GCN]{Graph Convolutional Network}
\acro{gnn}[GNN]{Graph Neural Network}
\acro{gpu}[GPU]{Graphics Processing Unit}
\acro{gru}[GRU]{Gated Recurrent Unit}
\acro{gtpt}[G2PT]{Gloss to Pose Transformer}
\acro{hmm}[HMM]{Hidden Markov Model}
\acro{hpe}[HPE]{Hand Pose Enhancer}
\acro{isl}[ISL]{Irish Sign Language}
\acro{lstm}[LSTM]{Long Short-Term Memory}
\acro{mha}[MHA]{Multi-Headed Attention}
\acro{mtc}[MTC]{Monocular Total Capture}
\acro{mse}[MSE]{Mean Squared Error}
\acro{mdn}[MDN]{Mixture Density Network}
\acro{moe}[MoE]{Mixture-of-Expert}
\acro{nmt}[NMT]{Neural Machine Translation}
\acro{nlp}[NLP]{Natural Language Processing}
\acro{ph12}[PHOENIX12]{RWTH-PHOENIX-Weather-2012}
\acro{ph14}[PHOENIX14]{RWTH-PHOENIX-Weather-2014}
\acro{ph14t}[PHOENIX14\textbf{T}]{RWTH-PHOENIX-Weather-2014\textbf{T}}
\acro{pof}[POF]{Part Orientation Field}
\acro{paf}[PAF]{Part Affinity Field}
\acro{pttt}[P2TT]{Pose to Text Transformer}
\acro{relu}[RELU]{Rectified Linear Units}
\acro{rnn}[RNN]{Recurrent Neural Network}
\acro{rouge}[ROUGE]{Recall-Oriented Understudy for Gisting Evaluation}
\acro{sgd}[SGD]{Stochastic Gradient Descent}
\acro{sla}[SLA]{Sign Language Assessment}
\acro{slr}[SLR]{Sign Language Recognition}
\acro{slt}[SLT]{Sign Language Translation}
\acro{slp}[SLP]{Sign Language Production}
\acro{smt}[SMT]{Statistical Machine Translation}
\acro{slva}[SLVA]{Sign Language Video Anonymisation}

\acro{ttgt}[T2GT]{Text to Gloss Transformer}
\acro{ttpt}[T2PT]{Text to Pose Transformer}
\acro{ttp}[T2P]{Text to Pose}
\acro{ttgtp}[T2G2P]{Text to Gloss to Pose}

\acro{vae}[VAE]{Variational Autoencoder}

\acro{wer}[WER]{Word Error Rate}

\end{acronym}
\usepackage{dblfloatfix}

\usepackage[dvipsnames]{xcolor}

\usepackage{multirow}

\newcommand{\R}{\mathbb{R}}
\newcommand{\methodNameFull}{Skeletal Graph Self-Attention}
\newcommand{\methodName}{$SGSA$}
\newcommand{\mc}{\mathcal}

\usepackage{enumitem,xcolor}
\newlist{coloritemize}{itemize}{1}
\setlist[coloritemize]{label=\textcolor{itemizecolor}{\textbullet},font=\bfseries\color{itemizecolor}}
\colorlet{itemizecolor}{red}

\usepackage{accents}

\usepackage[pagebackref,breaklinks,colorlinks]{hyperref}

\usepackage[capitalize]{cleveref}
\crefname{section}{Sec.}{Secs.}
\Crefname{section}{Section}{Sections}
\Crefname{table}{Table}{Tables}
\crefname{table}{Tab.}{Tabs.}

\begin{document}

\title{Skeletal Graph Self-Attention: Embedding a Skeleton Inductive Bias into Sign Language Production}

\author{Ben Saunders, Necati Cihan Camgoz, Richard Bowden\\
University of Surrey\\
{\tt\small \{b.saunders, n.camgoz, r.bowden\}@surrey.ac.uk}
}

\maketitle
\begin{abstract}
Recent approaches to \ac{slp} have adopted spoken language \ac{nmt} architectures, applied without sign-specific modifications. In addition, these works represent sign language as a sequence of skeleton pose vectors, projected to an abstract representation with no inherent skeletal structure.  

In this paper, we represent sign language sequences as a skeletal graph structure, with joints as nodes and both spatial and temporal connections as edges. To operate on this graphical structure, we propose \methodNameFull{} (\methodName{}), a novel graphical attention layer that embeds a skeleton inductive bias into the \ac{slp} model. Retaining the skeletal feature representation throughout, we directly apply a spatio-temporal adjacency matrix into the self-attention formulation. This provides structure and context to each skeletal joint that is not possible when using a non-graphical abstract representation, enabling fluid and expressive sign language production. 

We evaluate our \methodNameFull{} architecture on the challenging RWTH-PHOENIX-Weather-2014\textbf{T} (PHOENIX14\textbf{T}) dataset, achieving state-of-the-art back translation performance with an 8\% and 7\% improvement over competing methods for the dev and test sets.
\end{abstract}

\section{Introduction} \label{sec:intro}

Sign languages are rich visual languages, the native languages of the Deaf communities. Comprised of both manual (hands) and non-manual (face and body) features, sign languages can be visualised as spatio-temporal motion of the hands and body \cite{sutton1999linguistics}. When signing, the local context of motions is particularly important, such as the connections between fingers in a sign, or the lip patterns when mouthing \cite{pfau2010nonmanuals}. Although commonly represented via a graphical avatar, more recent deep learning approaches to \acf{slp} have represented sign as a continuous sequence of skeleton poses \cite{saunders2021continuous,stoll2018sign,zelinka2020neural}.

Due to the recent success of \acf{nmt}, computational sign language research often naively applies spoken language architectures without sign-specific modifications. However, the domains of sign and spoken language are drastically different \cite{stokoe1980sign}, with the continuous nature and inherent spatial structure of sign requiring sign-dependent architectures. Saunders \etal \cite{saunders2020progressive} introduced \textit{Progressive Transformers}, an \ac{slp} architecture specific to a continuous skeletal representation. However, this still projects the skeletal input to an abstract feature representation, losing the skeletal inductive bias inherent to the body, where each joint upholds its own spatial representation. Even if spatio-temporal skeletal relationships can be maintained in an latent representation, a trained model may not correctly learn this complex structure.

\begin{figure*}[t!]
    \centering
    \includegraphics[width=1.0\linewidth]{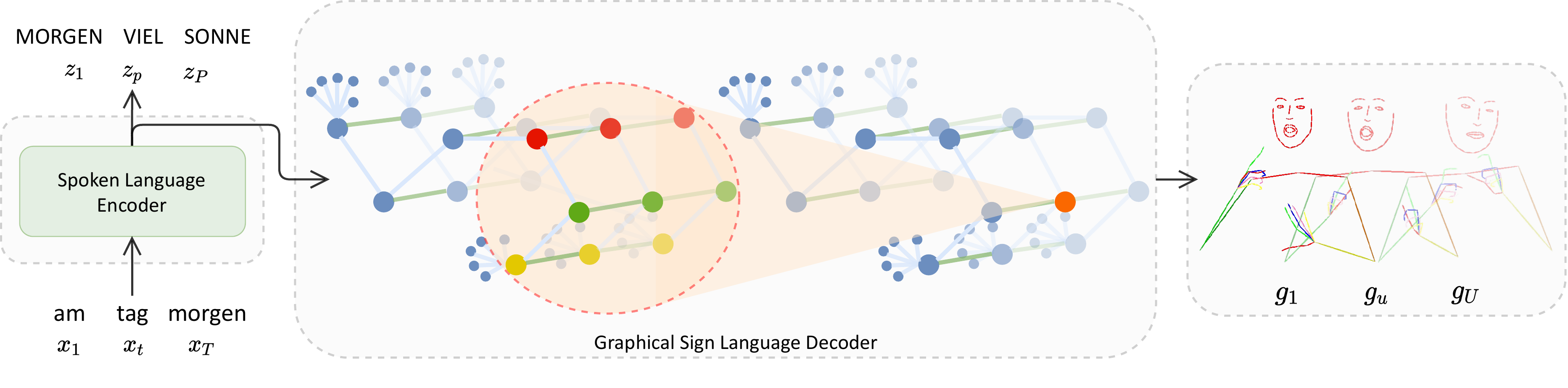}
    \caption{An overview of our proposed \ac{slp} network, showing an initial translation from a spoken language sentence using a text encoder, with gloss supervision. A subsequent skeletal graphical structure is formed, with multiple proposed \methodNameFull{} layers applied to embed a skeleton inductive bias and produce expressive sign language sequences.}
    \label{fig:skeletal_graph}
\end{figure*}%

Graphical structures can be used to represent pairwise relationships between objects in an ordered space. \acp{gnn} are neural models used to capture graphical relationships, and predominantly operate on a high-level graphical structure \cite{bruna2014spectral}, with each node containing an abstract feature representation and relationships occurring at the meta level. Conversely, skeleton pose sequences can be defined as spatio-temporal graphical representations, with both intra-frame spatial adjacency between limbs and inter-frame temporal adjacency between frames. In this work, we employ attention mechanisms as global graphical structures, with each node attending to all others. Even though there have been attempts to combine graphical representations and attention \cite{yun2019graph,dwivedi2020generalization,velivckovic2017graph}, there has been no work on graphical self-attention specific to a spatio-temporal skeletal structure.

In this paper, we represent sign language sequences as spatio-temporal skeletal graphs, the first \ac{slp} model to operate with a graphical structure. As seen in the centre of Figure \ref{fig:skeletal_graph}, we encode skeletal joints as nodes, $\mc{J}$ (blue dots), and natural limb connections as edges, $\mc{E}$, with both spatial (blue lines) and temporal (green lines) relationships. Operating on a graphical structure explicitly upholds the skeletal representation throughout, learning deeper and more informative features than using an abstract representation.

Additionally, we propose \methodNameFull{} (\methodName{}), a novel spatio-temporal graphical attention layer that embeds a hierarchical body inductive bias into the self-attention mechanism. We directly mask the self-attention by applying a sparse adjacency matrix to the weights of the value computation, ensuring a spatial information propagation. To the best of our knowledge, ours is the first work to embed a graphical structure directly into the self-attention mechanism. In addition, we expand our model to the spatio-temporal domain by modelling the temporal adjacency only on $\mc{N}$ neighbouring frames.

Our full \ac{slp} model can be seen in Figure \ref{fig:skeletal_graph}, initially translating from spoken language using a spoken language encoder with gloss supervision. The intermediary graphical structure is then processed by a graphical sign language decoder containing our proposed \methodName{} layers, with a final output of sign language sequences. We evaluate on the challenging \ac{ph14t} dataset, performing spatial and temporal ablation studies of the proposed \methodName{} architecture. Furthermore, we achieve state-of-the-art back translation results for the text to pose task, with an 8\% and 7\% performance increase over competing methods for the development and test sets respectively.

The contributions of this paper can be summarised as:
\begin{itemize} \itemsep0em
    \item The first \ac{slp} system to model sign language as a spatio-temporal graphical structure, applying both spatial and temporal adjacency.
    \item A novel \methodNameFull{} (\methodName{}) layer, that embeds a skeleton inductive bias into the model.
    \item State-of-the-art Text-to-Pose \ac{slp} results on the \ac{ph14t} dataset.
\end{itemize}

\section{Related Work} \label{sec:related_work}
\paragraph{Sign Language Production}
The past 30 years has seen extensive research into computational sign language \cite{liang1996sign,wilson1993neural}. Early work focused on isolated \ac{slr} \cite{lim2019isolated,grobel1997isolated}, with a subsequent move to continuous \ac{slr} \cite{camgoz2017subunets,koller2015continuous}. The task of \ac{slt} was introduced by Camgoz \etal \cite{camgoz2018neural} and has since become a prominent research area \cite{yin2020sign,camgoz2020multi,rodriguez2021important}. \acf{slp}, the automatic translation from spoken language sentences to sign language sequences, was initially tackled using avatar-based technologies \cite{lu2010collecting,elliott2008linguistic}. The rule-based \ac{smt} achieved partial success \cite{kayahan2019hybrid,kouremenos2018statistical}, albeit with costly, labour-intensive pre-processing.

Recently, there have been many deep learning approaches to \ac{slp} proposed \cite{zelinka2020neural,stoll2018sign,xiao2020skeleton,miyazaki2020machine,saunders2020everybody,saunders2021anonysign,saunders2021mixed,huang2021towards}, with Saunders \etal{} achieving state-of-the-art results with gloss supervision \cite{saunders2021mixed}. These works predominantly represent sign languages as sequences of skeletal frames, with each frame encoded as a vector of joint coordinates \cite{saunders2021continuous} that disregards any spatio-temporal structure available within a skeletal representation. In addition, these models apply standard spoken language architectures \cite{vaswani2017attention}, disregarding the structural format of the skeletal data. Conversely, in this work we propose a novel spatio-temporal graphical attention layer that injects an inductive skeletal bias into \ac{slp}.

\begin{figure*}[t!]
    \centering
    \includegraphics[width=0.6\linewidth]{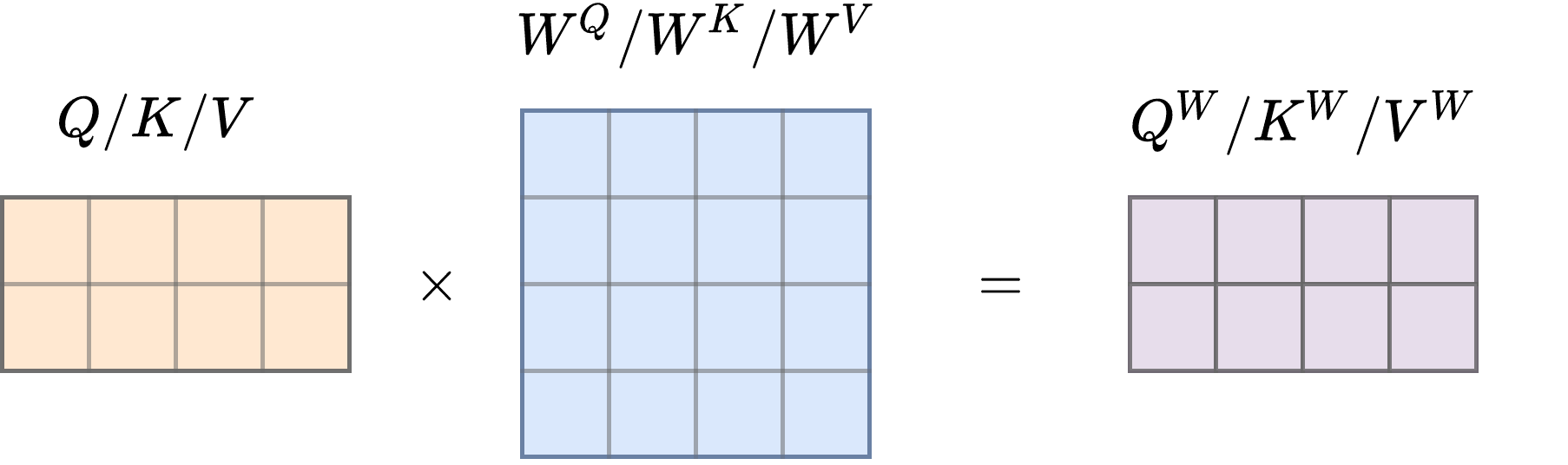}
    \caption{Weighted calculation of Queries, $Q$, Keys, $K$ and Values, $V$, for global self-attention.}
    \label{fig:self_attention}
\end{figure*}%

\paragraph{Graph Neural Networks}

A graph is a data structure consisting of nodes, $\mc{J}$, and edges, $\mc{E}$, where $\mc{E}$ defines the relationships between $\mc{J}$. \acfp{gnn} \cite{bruna2014spectral} apply neural layers on these graphical structures to learn representations \cite{zhou2020graph,qi2018learning}, classify nodes \cite{yan2018spatial,yao2019graph} or generate new data \cite{li2018learning,yang2019conditional}. A skeleton pose representation can be structured as a graph, with joints as $\mc{J}$ and natural limb connections as $\mc{E}$ \cite{straka2011skeletal,shi2019skeleton}. \acp{gnn} have been proposed for operating on such dynamic skeletal graphs, in the context of action recognition \cite{yan2018spatial,shi2019skeleton,kao2019graph,mao2019learning} and human pose estimation \cite{straka2011skeletal}.

Attention networks can be formalised as a fully connected \ac{gnn}, where the adjacency between each word, $\mc{E}$, is a weighting learnt using self-attention. Expanding this, \acp{gat} \cite{velivckovic2017graph} define explicit weighted adjacency between nodes, achieving state-of-the-art results across multiple domains \cite{kosaraju2019social,song2019session,busbridge2019relational}. Recently, there have been multiple graphical transformer architectures proposed \cite{yun2019graph,dwivedi2020generalization,koncel2019text,kreuzer2021rethinking}, which have been extended to the spatio-temporal domain for applications such as multiple object tracking \cite{chu2021transmot} and pedestrian tracking \cite{yu2020spatio}.

However, there has been no work on graphical attention mechanisms where the features of each time step holds a relevant graphical structure. We build a spatio-temporal graphical architecture that operates on a skeletal representation per frame, explicitly injecting a skeletal inductive bias into the model. There have been some applications of \acp{gnn} in computational sign language in the context of \ac{slr} \cite{de2019spatial,meng2021attention,flasinski2010use,tolba2012proposed,jiang2021skeleton}. We extend these works to the \ac{slp} domain with our proposed \methodNameFull{} architecture.
\paragraph{Local Attention}
Attention mechanisms have demonstrated strong \ac{nlp} performance \cite{bahdanau2015neural}, particularly with the introduction of transformers \cite{vaswani2017attention}. Although proposed with global context \cite{bahdanau2015neural}, more recent works have selectively restricted attention to only a local context \cite{yang2018modeling,child2019generating,liu2018generating} or the top-k tokens\cite{zhao2019explicit}, often due to computational issues or to enable long-range dependencies. In this paper, we propose using local attention to represent temporal adjacency within our graphical skeletal structure.

\section{Background} \label{sec:background}

In this section, we provide a brief background on self-attention. Attention mechanisms were initially proposed to overcome the information bottleneck found in encoder-decoder architectures \cite{bahdanau2015neural,luong2015effective}. Transformers \cite{vaswani2017attention} apply multiple scaled self-attention layers in both encoder and decoder modules, where the input is a set of queries, $Q \in \R^{d_{k}}$, and keys, $K \in \R^{d_{k}}$, and values, $V\in \R^{d_{v}}$. Self-attention aims to learn a context value for each time-step as a weighted sum of all values, where the weight is determined by the relationship of the query with each corresponding key. An associated weight vector, $W^{Q/K/V}$, is first applied to each input, as shown in Figure \ref{fig:self_attention}, as:
\begin{equation}
\label{eq:weighted_variables}
    Q^{W} = Q \cdot W^{Q}, \;\;\; K^{W} = K \cdot W^{K}, \;\;\; V^{W} = V \cdot W^{V}
\end{equation}
where $W^{Q} \in \R^{d_{model} \times d_{k}}$, $W^{K} \in \R^{d_{model} \times d_{k}}$ and $W^{V} \in \R^{d_{model} \times d_{v}}$ are weights related to each input variable and $d_{model}$ is the dimensionality of the self-attention layer. Formally, scaled self-attention ($\textrm{SA}$) outputs a weighted vector combination of values, $V^{W}$, by the relevant queries, $Q^{W}$, keys, $K^{W}$, and dimensionality, $d_{k}$, as:
\begin{equation}
\label{eq:self-attention}
    \textrm{SA} (Q,K, V) = 
    \text{softmax} (\frac{Q^{W}  (K^{W})^{T}}{\sqrt{d_{k}}})V^{W}
\end{equation}

\ac{mha} applies $h$ parallel attention mechanisms to the same input queries, keys and values, each with different learnt parameters. In the initial architecture \cite{vaswani2017attention}, the dimensionality of each head is proportionally smaller than the full model, $d_{h} = d_{model} / h$. The output of each head is then concatenated and projected forward, as:
\begin{align}
\label{eq:multi_head_attention}
    \textrm{MHA}(Q,K, & V)  =  [head_{1}, ... ,head_{h}] \cdot W^{O}, \nonumber \\
      &  \textrm{where} \medspace head_{i} = \textrm{SA}(Q^{W}, K^{W}, V^{W})
\end{align}
where $W^{O} \in \R^{d_{model} \times d_{model}}$. In this paper, we introduce \methodNameFull{} layers that inject a skeletal inductive bias into the self-attention mechanism.

\section{Methodology} \label{sec:methodology}

\begin{figure*}[t!]
    \centering
    \includegraphics[width=0.99\linewidth]{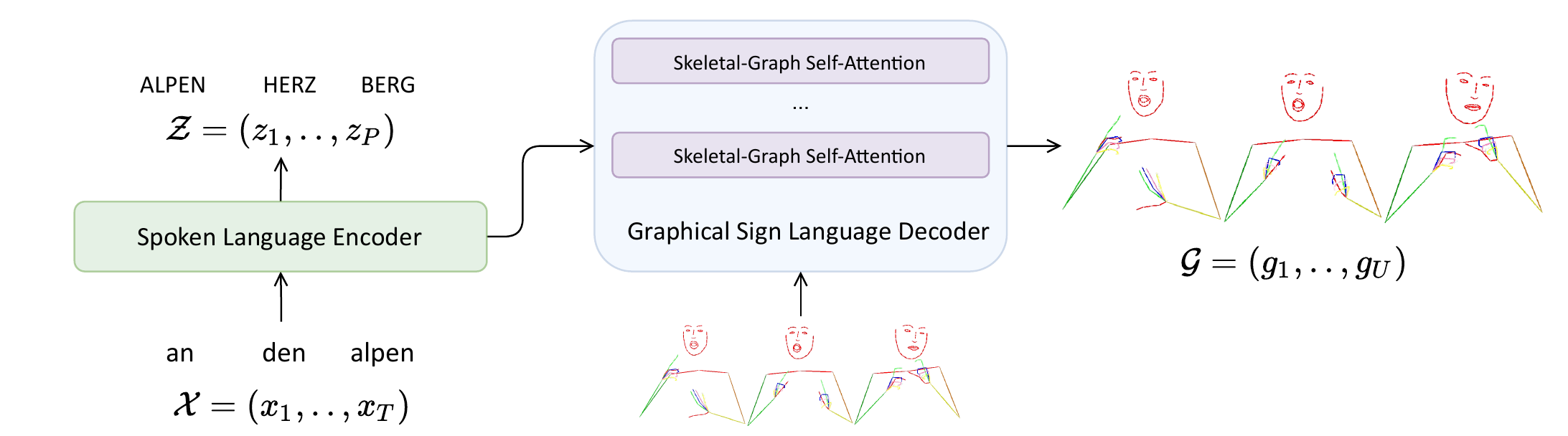}
    \caption{Overview of the proposed model architecture, detailing the Spoken Language Encoder (Sec. \ref{sec:language_encoder}) and the Graphical Sign Language Decoder (Sec. \ref{sec:sign_decoder}). We propose novel \methodNameFull{} layers to operate on the sign language skeletal graphs, $\mc{G}$.}
    \label{fig:overview}
\end{figure*}%

The ultimate goal of \ac{slp} is to automatically translate from a source spoken language sentence, \hbox{$\mc{X} = (x_1, ..., x_T)$} with $\mc{T}$ words, to a target sign language sequence, \hbox{$\mc{G} = (g_1, ..., g_U)$} of $\mc{U}$ time steps. Additionally, an intermediary gloss\footnote{Glosses are a written representation of sign, defined as minimal lexical items.} sequence representation can be used, \hbox{$\mc{Z} = (z_{1},...,z_{P})$} with $P$ glosses. Current approaches \cite{saunders2021continuous,stoll2018sign,zelinka2020neural} predominantly represent sign language as a sequence of skeletal frames, with each frame containing a vector of body joint coordinates. In addition, they project this skeletal structure to an abstract representation before being processed by the model \cite{saunders2020progressive}. However, this approach removes all spatial information contained within the skeletal data, restricting the model to only learning the internal relationships within a latent representation.

Contrary to previous work, in this paper we represent sign language sequences as spatio-temporal skeletal graphs, $\mc{G}$, as in the centre of Figure \ref{fig:skeletal_graph}. As per graph theory \cite{bollobas2013modern}, $\mc{G}$ can be formulated as a function of nodes, $\mc{J}$ and edges, $\mc{E}$. We define $\mc{J}$ as the skeleton pose sequence of temporal length $\mc{U}$ and spatial width $\mc{S}$, with each node representing a single skeletal joint coordinate from a single frame (blue dots in Fig. \ref{fig:skeletal_graph}). $\mc{S}$ is therefore the dimensionality of the skeleton representation of each frame. $\mc{E}$ can be represented as a spatial adjacency matrix, $\mc{A}$, defined as the natural limb connections between skeleton joints both of its own frame (blue lines) and of neighbouring frames (green lines). 

As outlined in Sec. \ref{sec:background}, classical self-attention operates with global context over all sequence time-steps. However, a skeletal inductive bias can be embedded into a model by restricting attention to only the natural limb connections within the skeleton. To embed a skeleton inductive bias into self-attention, we propose a novel \methodNameFull{} (\methodName{}) layer that operates with sparse attention. Modeled within a transformer decoder, \methodName{} retains the original skeletal structure throughout multiple deep layers, ensuring the processing of spatio-temporal information contained in skeletal pose sequences. In-built adjacency matrices of both intra- and inter-frame relationships provide structure and context directly to each skeletal joint that is not possible when using a non-graphical abstract representation. 

In the rest of this section, we outline the full \ac{slp} model, containing a spoken language encoder and a graphical sign language decoder, with an overview shown in Figure \ref{fig:overview}.

\subsection{Spoken Language Encoder} \label{sec:language_encoder}

As shown on the left of Figure \ref{fig:overview}, we first translate from a spoken language sentence, $\mc{X}$, of dimension $\mc{E} \times \mc{T}$, where $\mc{E}$ is the encoder embedding size, to a sign language representation, $\mc{R} = (r_1, ..., r_U)$ (Fig. \ref{fig:skeletal_graph} Left). We build a classical transformer encoder \cite{vaswani2017attention} that applies self-attention using the global context of a spoken language sequence. $\mc{R}$ is represented with a spatio-temporal structure, containing identical temporal length, $\mc{U}$, and spatial shape, $\mc{S}$, as the final skeletal graph, $\mc{G}$. This structure enables a graphical processing by the proposed sign language decoder. Additionally, as proposed in \cite{saunders2021mixed}, we employ a gloss supervision  to the intermediate sign language representation. This prompts the model to learn a meaningful latent sign representation for the ultimate goal of sign language production.

\subsection{Graphical Sign Language Decoder} \label{sec:sign_decoder}

Given the intermediary sign language representation, $\mc{R} \in $, we build an auto-regressive transformer decoder containing our novel Skeletal Graph Self-Attention (SGSA) layers (Figure \ref{fig:overview} middle). This produces a graphical sign language sequence, $\mc{\hat{G}}$, of spatial shape, $\mc{S}$, and temporal length, $\mc{U}$.

\paragraph{Spatial Adjacency}

\begin{figure*}[t!]
    \centering
    \includegraphics[width=0.85\linewidth]{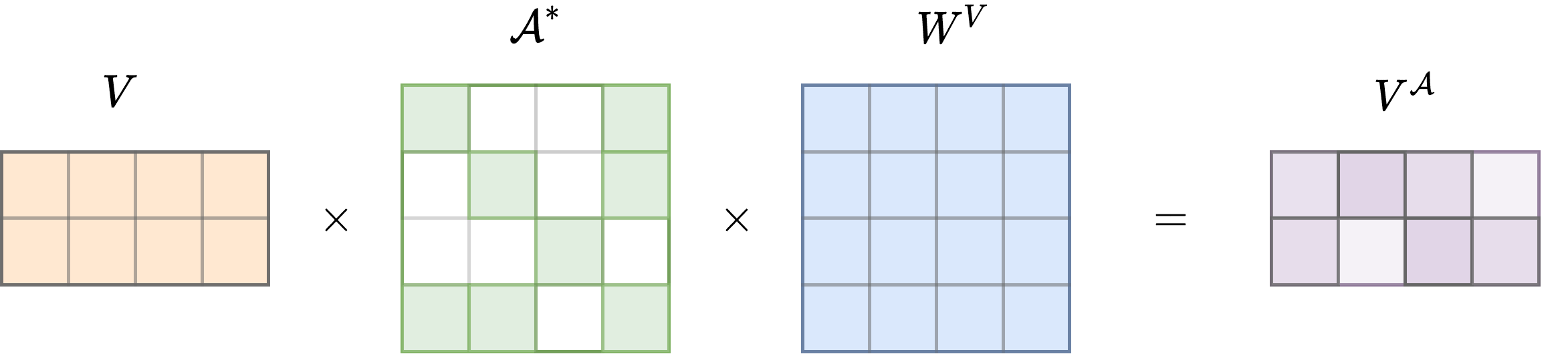}
    \caption{Skeletal Graph Self-Attention: Weighted calculation of Values, $V$, masked with a spatio-temporal adjacency matrix $\mc{A}^{*}$ to embed a skeleton inductive bias.}
    \label{fig:graph_self_attention}
\end{figure*}

We define a spatial adjacency matrix, $\mc{A} \in \R^{\mc{S} \times \mc{S}}$, expressed as a sparse attention map, as seen in Figure \ref{fig:graph_self_attention}. $\mc{A}$ contains a spatial skeleton adjacency structure, modelled as the natural skeletal limb connections within a frame (blue lines in Fig. \ref{fig:skeletal_graph}). $\mc{A}$ can be formalised as:
\begin{equation} \label{eq:A_spatial}
    \mc{A}_{i,j} = 
    \begin{cases}
    1,  & \textit{if } \textrm{Con}(i,j) \\
    0,  & \textit{otherwise}
    \end{cases}
\end{equation}
where $\textrm{Con}(i,j) = True$ if joints $i$ and $j$ are connected. For example, the skeletal elbow joint is connected to the skeletal wrist joint. We use an undirected graph representation, defining $\mc{E}$ as bidirectional edges.

\paragraph{Temporal Adjacency}

We expand the spatial adjacency matrix to the spatio-temporal domain by modelling the inter-frame edges of the skeletal graph structure (green lines in Fig. \ref{fig:skeletal_graph}). The updated spatial-temporal adjacency matrix can be formalised as $\mc{A} \in \R^{\mc{S} \times \mc{S} \times \mc{U}}$. We set $\mc{N}$ as the temporal distance that defines `adjacent', where edges are established as both same joint connections and natural limb connections between the $\mc{N}$ adjacent frames. In the standard attention shown in Sec. \ref{sec:background}, each time-step can globally attend to all others, which can be modelled as $\mc{N} = \infty$. We formalise our spatio-temporal adjacency matrix, as:
\begin{equation} \label{eq:A_spatial_temporal}
    \mc{A}_{i,j,t} = 
    \begin{cases}
    1,  & \textit{if } \textrm{Con}(i,j) \textit{ and } t \leq \mc{N} \\
    0,  & \textit{otherwise}
    \end{cases}
\end{equation}
where $t$ is the temporal distance from the reference frame, $t = u_{\text{}} - u_{\text{ref}}$.

\begin{table*}[t]
\centering
\resizebox{0.85\linewidth}{!}{%
\begin{tabular}{@{}p{2.8cm}ccccc|ccccc@{}}
\toprule
  \multicolumn{1}{c}{Skeletal Graph}     & \multicolumn{5}{c}{DEV SET}  & \multicolumn{5}{c}{TEST SET} \\ 
\multicolumn{1}{c|}{Layers, $\mc{L}$:}  & BLEU-4         & BLEU-3         & BLEU-2         & BLEU-1         & ROUGE          & BLEU-4         & BLEU-3         & BLEU-2         & BLEU-1         & ROUGE          \\ \midrule
\multicolumn{1}{c|}{0 (4 SA)} & 14.25 & 17.73 & 23.47 & 34.79 & 37.65 & 13.64 & 17.03 & 23.09 & 35.03 & 36.59   \\ \hline
\multicolumn{1}{c|}{1} & 14.37 & 17.67 & 23.13 & 33.95 & 36.98 & 13.63 & 17.08 & 23.17 & 35.39 & 37.05 \\
\multicolumn{1}{c|}{2} & 14.50 & 18.14 & 24.10 & 35.96 & 38.09 & 13.85 & 17.23 & 23.14 & 34.93 & 37.33\\
\multicolumn{1}{c|}{3} & 14.53 & 18.02 & 24.00 & 35.71 & 37.62 & 13.72 & 17.23 & 23.10 & 34.45 & 36.99 \\
\multicolumn{1}{c|}{4} & 14.68 & 18.30 & \textbf{24.31} & \textbf{36.16} & 38.51 & 14.05 & 17.59 & 23.73 & 35.63 & 37.47 \\
\multicolumn{1}{c|}{5} & \textbf{14.72} & \textbf{18.39} & 24.29 & 35.79 & \textbf{38.72} & \textbf{14.27} & \textbf{17.79} & \textbf{23.79} & \textbf{35.72} & \textbf{37.79} \\
\bottomrule
\end{tabular}%
}
\caption{Impact of \methodNameFull{} layers, $\mc{L}$, on model performance.}
\label{tab:spatial_adjacency}
\end{table*}

\paragraph{Self-loops and Normalisation}

To account for information propagation loops back to the same joint \cite{bollobas2013modern}, we add self-loops to $\mc{A}$ using the identity matrix, $\mc{I} \in \R^{\mc{S} \times \mc{S}}$. In practice, due to our multi-dimensional skeletal representation, we add self-loops from each coordinate of the joint both to itself and all other coordinates of the same joint, which we define as $\mc{I}^{*} \in \R^{\mc{S} \times \mc{S}}$. Furthermore, to prevent numerical instabilities and exploding gradients \cite{bollobas2013modern}, we normalise the adjacency matrix by inversely applying the degree matrix, $\mc{D} \in \R^{\mc{S}}$. $\mc{D}$ is defined as the numbers of edges a node is connected to. Normalisation can be formulated as:
\begin{equation}
\label{eq:normalised_adjacency}
    \mc{A}^{*} = \mc{D}^{-1} (\mc{A} + \mc{I}^{*})
\end{equation}
where $\mc{A}^{*}$ is the normalised adjacency matrix. 

\paragraph{Skeletal Graph Self-Attention}

We apply $\mc{A}^{*}$ as a sparsely weighted mask onto the weighted value calculation, $V^{W} = V \cdot W^{V}$, of Eq. \ref{eq:weighted_variables}, ensuring that the values used in the weighted context for each node is only impacted by the adjacent nodes of the previous layer: 
\begin{equation}
\label{eq:adjacency_weighted_values}
    V^{\mc{A}} = V \cdot \mc{A}^{*} \cdot W^{V}
\end{equation}
where Figure \ref{fig:graph_self_attention} shows a visual representation of the sparse adjacent matrix $\mc{A}^{*}$ containing spatio-temporal connections, applied as a mask to the weighted calculation. With a value matrix containing a skeletal structure, $V\in \R^{\mc{S}}$, $\mc{A}^{*}$ restricts the information propagation of self-attention layers only through the spatial and temporal skeletal edges, $\mc{E}$, and thus embeds a skeleton inductive bias into the attention mechanism. 

We formally define a \methodNameFull{} (\methodName{}) layer by plugging both the weighted variable computation of Eq. \ref{eq:weighted_variables} and the adjacent weighted computation of Eq. \ref{eq:adjacency_weighted_values} into the self-attention Eq. \ref{eq:self-attention}, as:
\begin{align}
\label{eq:graph_self_attention}
    SGSA( Q,K & ,V ,A) = \nonumber \\
    \text{softmax} & (\frac{Q \cdot W^{Q} (K \cdot W^{K})^{T}}{\sqrt{d_{k}}})V \cdot \mc{A}^{*} \cdot W^{V}
\end{align}
where $d_{model} = \mc{S}$. This explicitly retains the spatial skeletal shape, $\mc{S}$, throughout the sign language decoder, enabling a spatial structure to be extracted.

To extend this to a multi-headed transformer decoder, we replace self-attention in Eq. \ref{eq:multi_head_attention} with our proposed \methodName{} layers. To retain the spatial skeletal representation within each head, the dimensionality of each head is kept as the full model dimension, $d_{h} = d_{model} = \mc{S}$, with the final projection layer enlarged to $h \times \mc{S}$.

We build our auto-regressive sign language decoder with $\mc{L}$ multi-headed \methodName{} sub-layers, interleaved with fully-connected layers and a final feed-forward layer, each with a consistent spatial dimension of $\mc{S}$. A residual connection \cite{he2016deep} and subsequent layer norm \cite{ba2016layer} is employed around each of the sub-layers, to aid training. As shown on the right of Figure \ref{fig:overview}, the final output of our sign language decoder module is a graphical skeletal sequence, $\hat{\mc{G}}$, that contains $\mc{U}$ frames of skeleton pose, each with a spatial shape of $\mc{S}$.

We train our sign language decoder using the \ac{mse} loss between the predicted sequence, $\mc{\hat{G}}$, and the ground truth sequence, $\mc{G}^{*}$. This is formalised as $\mathcal{L}_{\mathcal{MSE}} = \frac{1}{U} \sum_{i=1}^{u}  \hat{g}_{1:U} - {g^{*}}_{1:U} ) ^{2}$, where $\hat{g}$ and $g^{*}$ represent the frames of the produced and ground truth sign language sequences, respectively. We train our full \ac{slp} model end-to-end with a weighted combination of the encoder gloss supervision \cite{saunders2021mixed} and decoder skeleton pose losses.

\subsection{Sign Language Output}

Generating a sign language video from the produced graphical skeletal sequence, $\hat{\mc{G}}$, is then a trivial task, animating each frame in temporal order. Frame animation is done by connecting the nodes, $\mc{J}$, using the natural limb connections defined by $\mc{E}$, as seen in Fig. \ref{fig:skeletal_graph}.

\section{Experiments} \label{sec:experiments}

\paragraph{Dataset} \label{sec:dataset}
We evaluate our approach on the \ac{ph14t} dataset introduced by Camgoz et al. \cite{camgoz2018neural}, containing parallel sequences of 8257 German sentences, sign gloss translations and sign language videos. Other available sign datasets are either simple sentence repetition tasks of non-natural signing not appropriate for translation \cite{zhang2016chinese,efthimiou2007gslc}, or contain larger domains of discourse that currently prove difficult for the SLP field \cite{camgoz2021content4all,hanke2010dgs}. We extract 3D skeletal joint positions from the sign language videos to represent our spatio-temporal graphical skeletal structure. Manual and non-manual features of each video are first extracted in 2D using OpenPose \cite{cao2018openpose}, with the manuals lifted to 3D using the skeletal model estimation model proposed in \cite{zelinka2020neural}. We normalise the skeleton pose and set the spatial skeleton shape, $\mc{S}$, as 291, with 290 joint coordinates and 1 counter decoding value (as in \cite{saunders2020progressive}). Adjacency information, $\mc{A}$, is defined as the natural limb connections of 3D body, hand and face joints, as in \cite{zelinka2020neural}, where each coordinate of a joint is adjacent to both the coordinates of its own joint and all connected joints. We define the counter value as global adjacency, with connections to all joints.

\paragraph{Implementation Details}

\begin{table*}[t]
\centering
\resizebox{0.85\linewidth}{!}{%
\begin{tabular}{@{}p{2.8cm}ccccc|ccccc@{}}
\toprule
  \multicolumn{1}{c}{Temporal}  & \multicolumn{5}{c}{DEV SET}  & \multicolumn{5}{c}{TEST SET} \\ 
\multicolumn{1}{c|}{Adjacency, $\mc{N}$:}  & BLEU-4         & BLEU-3         & BLEU-2         & BLEU-1         & ROUGE          & BLEU-4         & BLEU-3         & BLEU-2         & BLEU-1         & ROUGE          \\ \midrule
\multicolumn{1}{c|}{$\infty$} & 14.72 & 18.39 & 24.29 & 35.79 & 38.72 & 14.27 & 17.79 & 23.79 & 35.72 & 37.79 \\
\multicolumn{1}{c|}{1} & \textbf{15.15} & 18.67 & 24.47 & 35.88 & 38.44 & \textbf{14.33} & 17.77 & 23.72 & 35.26 & 37.96 \\
\multicolumn{1}{c|}{2} & 15.09 & 18.51 & 24.43 & 36.17 & 38.04 & 14.07 & 17.62 & 23.91 & \textbf{36.28} & 37.82 \\
\multicolumn{1}{c|}{3} & 15.08 & \textbf{18.84} & 24.89 & 36.66 & 38.95 & 14.32 & \textbf{17.95} & \textbf{24.04} & 36.10 & \textbf{38.38} \\
\multicolumn{1}{c|}{5} & 14.90 & 18.81 & \textbf{25.30} & \textbf{37.31} & \textbf{39.55} & 14.21 & 17.79 & 23.98 & 35.88 & 38.44 \\
\bottomrule
\end{tabular}%
}
\caption{Impact of Temporal Adjacency, $\mc{N}$, on \methodName{} model performance}
\label{tab:temporal_adjacency}
\end{table*}

\begin{table*}[b!]
\centering
\resizebox{1.0\linewidth}{!}{%
\begin{tabular}{@{}p{2.8cm}ccccc|ccccc@{}}
\toprule
     & \multicolumn{5}{c}{DEV SET}  & \multicolumn{5}{c}{TEST SET} \\ 
\multicolumn{1}{c|}{Approach:}  & BLEU-4         & BLEU-3         & BLEU-2         & BLEU-1         & ROUGE          & BLEU-4         & BLEU-3         & BLEU-2         & BLEU-1         & ROUGE          \\ \midrule
\multicolumn{1}{r|}{Progressive Transformers \cite{saunders2020progressive}}    & 11.82 & 14.80 & 19.97 & 31.41 & 33.18 & 10.51 & 13.54 & 19.04 & 31.36 & 32.46 \\ 
\multicolumn{1}{r|}{Adversarial Training \cite{saunders2020adversarial}} & 12.65 & 15.61 & 20.58 & 31.84 & 33.68 & 10.81 & 13.72 & 18.99 & 30.93 & 32.74 \\
\multicolumn{1}{r|}{Mixture Density Networks \cite{saunders2021continuous}} & 11.54 & 14.48 & 19.63 & 30.94 & 33.40 & 11.68 & 14.55 & 19.70 & 31.56 & 33.19 \\ 
\multicolumn{1}{r|}{Mixture of Motion Primitives \cite{saunders2021mixed}} & 14.03 & 17.50 & 23.49 & 35.23 & 37.76 & 13.30 & 16.86 & 23.27 & \textbf{35.89} & 36.77 \\
\multicolumn{1}{r|}{\textbf{\methodNameFull{}}} & \textbf{15.15} & \textbf{18.67} & \textbf{24.47} & \textbf{35.88} & \textbf{38.44} & \textbf{14.33} & \textbf{17.77} & \textbf{23.72} & 35.26 & \textbf{37.96}  \\
\bottomrule
\end{tabular}%
}
\caption{Baseline comparisons on the \ac{ph14t} dataset for the \textit{Text to Pose} task.}
\label{tab:baselines}
\end{table*}

We setup our \ac{slp} model with a spoken language encoder of 2 layers, 4 heads and an embedding size, $\mc{E}$, of 256, and a graphical sign language decoder of 5 layers, 4 heads and an embedding size of $\mc{S}$. Our best performing model contains 9M trainable parameters. As proposed by Saunders \etal \cite{saunders2020progressive}, we apply Gaussian noise augmentation with a noise rate of 5. We train all parts of our network with Xavier initialisation \cite{glorot2010understanding}, Adam optimization \cite{kingma2014adam} with default parameters and a learning rate of $10^{-3}$. Our code is based on Kreutzer et al.'s NMT toolkit, JoeyNMT \cite{JoeyNMT}, and implemented using PyTorch \cite{paszke2017automatic}. 

\paragraph{Evaluation}

We use the back translation metric \cite{saunders2020progressive} for evaluation, which employs a pre-trained \ac{slt} model \cite{camgoz2020sign} to translate the produced sign pose sequences back to spoken language. We compute BLEU and ROUGE scores against the original input, with BLEU n-grams from 1 to 4 provided. The \ac{slp} evaluation protocols on the \ac{ph14t} dataset have been set by \cite{saunders2020progressive}. We share results on the \textit{Text to Pose (T2P)} task which constitutes the production of sign language sequences directly from spoken language sentences, the ultimate goal of an \ac{slp} system. We omit Gloss to Pose evaluation to focus on the more important spoken language translation task.

\paragraph{\methodNameFull{} Layers} \label{sec:quant}

We start our experiments on the proposed \methodNameFull{} layers, evaluating the effect of stacking multiple \methodName{} layers, $\mc{L}$, each with a multi-head size, $h$, of 4. We first ablate the effect of using no \methodName{} layers, and replacing them with 4 standard self-attention layers, as described in Section \ref{sec:background}. We then build our graphical sign language decoder with 1 to 5 \methodName{} layers, with each model retaining a constant spoken language encoder size and a global temporal adjacency.

Table \ref{tab:spatial_adjacency} shows that using standard self-attention layers achieves the worst performance of 14.25 BLEU-4, showing the benefit of our proposed \methodName{} layers. Increasing the number of \methodName{} layers, as expected, increases model performance to a peak of 14.72 BLEU-4. A larger number of layers enables a deeper representation of the skeletal graph and thus provides a stronger skeleton inductive bias to the model.  In lieu of this, for the rest of our experiments we build our sign language decoder with five \methodName{} layers.

\paragraph{Temporal Adjacency}
In our next set of experiments, we examine the impact of the temporal adjacency distance, $\mc{N}$, defined in Sec. \ref{sec:sign_decoder}. In order to logically set $\mc{N}$, we analyse the trained temporal attention matrix of the best performing decoder evaluated above. We notice that the attention predominantly falls on the last 3 frames, as the model learns to attend to the local temporal context of skeletal motion. Manually restricting the temporal attention provides this information as an inductive bias into the model, rather than relying on this being learnt. 

Table \ref{tab:temporal_adjacency} shows results of our temporal adjacency evaluation, ranging from an infinite adjacency (no constraint) to $\mc{N} \in [1,5]$. It can be seen that a temporal adjacency distance of one achieves the best BLEU-4 performance. Note: Although we report BLEU of n-grams 1-4 for completeness, we use BLEU-4 as our final evaluation metric to enable a clear result. Although counter-intuitive to the global self-attention utilised by a transformer decoder, we believe this is modelling the Markov property, where future frames only depend on the current state. Due to the intermediary gloss supervision \cite{saunders2021mixed}, the defined sign language representation, $\mc{R}$, should contain all frame-level information relevant to a sign language translation. The sign language decoder then has the sole task of accurately animating each skeletal frame. Therefore, a single temporal adjacency in the graphical decoder makes sense, as no new information is required to be learnt from temporally distant frames.

\paragraph{Baseline Comparisons}

We compare the performance of the proposed \methodNameFull{} architecture against 4 baseline \ac{slp} models: 1) Progressive transformers \cite{saunders2020progressive}, which applied the classical transformer architecture to sign language production. 2) Adversarial training \cite{saunders2020adversarial}, which utilised an adversarial discriminator to prompt more expressive productions, 3) \acp{mdn} \cite{saunders2021continuous}, which modelled the variation found in sign language using multiple distributions to parameterise the entire prediction subspace, and 4) Mixture of Motion Primitives (\textsc{MOMP}) \cite{saunders2021mixed}, which split the \ac{slp} task into two distinct jointly-trained sub-tasks and learnt a set of motion primitives for animation.

Table \ref{tab:baselines} presents \textit{Text to Pose} results, showing that \methodName{} achieves 15.15/14.33 BLEU-4 for the development and test sets respectively, an 8/7\% improvement over the state-of-the-art. These results highlight the significant success of our proposed \methodName{} layers. We have shown that representing sign pose skeletons in a graphical skeletal structure and embedding a skeletal inductive bias into the self-attention mechanism enables a fluid and expressive sign language production.

\section{Conclusion} \label{sec:conclusion}
In this paper, we proposed a skeletal graph structure for \ac{slp}, with joints as nodes and both spatial and temporal connections as edges. We proposed a novel graphical attention layer, \methodNameFull{}, to operate on the graphical skeletal structure. Retaining the skeletal feature representation throughout, we directly applied a spatio-temporal adjacency matrix into the self-attention formulation, embedding a skeleton inductive bias for expressive sign language production. We evaluated \methodName{} on the challenging \ac{ph14t} dataset, achieving state-of-the-art back translation performance with an 8\% and 7\% improvement over competing methods for the dev and test set. For future work, we aim to apply \methodName{} layers to the wider computational sign language tasks of \ac{slr} and \ac{slt}.

\section{Acknowledgements}

This work received funding from the SNSF Sinergia project `SMILE' (CRSII2 160811), the European Union's Horizon2020 research and innovation programme under grant agreement no. 762021 `Content4All' and the EPSRC project `ExTOL' (EP/R03298X/1). This work reflects only the authors view and the Commission is not responsible for any use that may be made of the information it contains.

\maketitle

{\small
\bibliographystyle{ieee_fullname}
\bibliography{bibliography}

\begin{thebibliography}{10}\itemsep=-1pt

\bibitem{ba2016layer}
Jimmy~Lei Ba, Jamie~Ryan Kiros, and Geoffrey~E Hinton.
\newblock {Layer Normalization}.
\newblock {\em arXiv preprint arXiv:1607.06450}, 2016.

\bibitem{bahdanau2015neural}
Dzmitry Bahdanau, Kyunghyun Cho, and Yoshua Bengio.
\newblock {Neural Machine Translation by Jointly Learning to Align and
  Translate}.
\newblock {\em Proceedings of the International Conference on Learning
  Representations (ICLR)}, 2015.

\bibitem{bollobas2013modern}
B{\'e}la Bollob{\'a}s.
\newblock {\em {Modern graph theory}}.
\newblock Springer Science \& Business Media, 2013.

\bibitem{bruna2014spectral}
Joan Bruna, Wojciech Zaremba, Arthur Szlam, and Yann LeCun.
\newblock {Spectral Networks and Locally Connected Networks on Graphs}.
\newblock In {\em Proceedings of the International Conference on Learning
  Representations (ICLR)}, 2014.

\bibitem{busbridge2019relational}
Dan Busbridge, Dane Sherburn, Pietro Cavallo, and Nils~Y Hammerla.
\newblock {Relational Graph Attention Networks}.
\newblock {\em arXiv preprint arXiv:1904.05811}, 2019.

\bibitem{camgoz2017subunets}
Necati~Cihan Camgoz, Simon Hadfield, Oscar Koller, and Richard Bowden.
\newblock {SubUNets: End-to-end Hand Shape and Continuous Sign Language
  Recognition}.
\newblock In {\em Proceedings of the IEEE International Conference on Computer
  Vision (ICCV)}, 2017.

\bibitem{camgoz2018neural}
Necati~Cihan Camgoz, Simon Hadfield, Oscar Koller, Hermann Ney, and Richard
  Bowden.
\newblock {Neural Sign Language Translation}.
\newblock In {\em Proceedings of the IEEE Conference on Computer Vision and
  Pattern Recognition (CVPR)}, 2018.

\bibitem{camgoz2020multi}
Necati~Cihan Camgoz, Oscar Koller, Simon Hadfield, and Richard Bowden.
\newblock {Multi-channel Transformers for Multi-articulatory Sign Language
  Translation}.
\newblock In {\em Assistive Computer Vision and Robotics Workshop (ACVR)},
  2020.

\bibitem{camgoz2020sign}
Necati~Cihan Camgoz, Oscar Koller, Simon Hadfield, and Richard Bowden.
\newblock {Sign Language Transformers: Joint End-to-end Sign Language
  Recognition and Translation}.
\newblock In {\em Proceedings of the IEEE Conference on Computer Vision and
  Pattern Recognition (CVPR)}, 2020.

\bibitem{camgoz2021content4all}
Necati~Cihan Camgoz, Ben Saunders, Guillaume Rochette, Marco Giovanelli,
  Giacomo Inches, Robin Nachtrab-Ribback, and Richard Bowden.
\newblock {Content4All Open Research Sign Language Translation Datasets}.
\newblock In {\em IEEE International Conference on Automatic Face and Gesture
  Recognition (FG)}, 2021.

\bibitem{cao2018openpose}
Zhe Cao, Gines Hidalgo, Tomas Simon, Shih-En Wei, and Yaser Sheikh.
\newblock {OpenPose: Realtime Multi-Person 2D Pose Estimation using Part
  Affinity Fields}.
\newblock In {\em Proceedings of the IEEE Conference on Computer Vision and
  Pattern Recognition (CVPR)}, 2017.

\bibitem{child2019generating}
Rewon Child, Scott Gray, Alec Radford, and Ilya Sutskever.
\newblock {Generating Long Sequences with Sparse Transformers}.
\newblock {\em arXiv preprint arXiv:1904.10509}, 2019.

\bibitem{chu2021transmot}
Peng Chu, Jiang Wang, Quanzeng You, Haibin Ling, and Zicheng Liu.
\newblock {TransMOT: Spatial-Temporal Graph Transformer for Multiple Object
  Tracking}.
\newblock {\em arXiv preprint arXiv:2104.00194}, 2021.

\bibitem{de2019spatial}
Cleison~Correia de Amorim, David Mac{\^e}do, and Cleber Zanchettin.
\newblock {Spatial-Temporal Graph Convolutional Networks for Sign Language
  Recognition}.
\newblock In {\em International Conference on Artificial Neural Networks},
  2019.

\bibitem{dwivedi2020generalization}
Vijay~Prakash Dwivedi and Xavier Bresson.
\newblock {A Generalization of Transformer Networks to Graphs}.
\newblock {\em arXiv preprint arXiv:2012.09699}, 2020.

\bibitem{efthimiou2007gslc}
Eleni Efthimiou and Stavroula-Evita Fotinea.
\newblock {GSLC: Creation and Annotation of a Greek Sign Language Corpus for
  HCI}.
\newblock In {\em International Conference on Universal Access in
  Human-Computer Interaction}, 2007.

\bibitem{elliott2008linguistic}
Ralph Elliott, John~RW Glauert, JR Kennaway, Ian Marshall, and Eva Safar.
\newblock {Linguistic Modelling and Language-Processing Technologies for
  Avatar-based Sign Language Presentation}.
\newblock {\em Universal Access in the Information Society}, 2008.

\bibitem{flasinski2010use}
Mariusz Flasi{\'n}ski and Szymon My{\'s}li{\'n}ski.
\newblock {On The Use of Graph Parsing for Recognition of Isolated Hand
  Postures of Polish Sign Language}.
\newblock {\em Pattern Recognition}, 2010.

\bibitem{glorot2010understanding}
Xavier Glorot and Yoshua Bengio.
\newblock {Understanding the Difficulty of Training Deep Feedforward Neural
  Networks}.
\newblock In {\em Proceedings of the International Conference on Artificial
  Intelligence and Statistics (AISTATS)}, 2010.

\bibitem{grobel1997isolated}
Kirsti Grobel and Marcell Assan.
\newblock {Isolated Sign Language Recognition using Hidden Markov Models}.
\newblock In {\em IEEE International Conference on Systems, Man, and
  Cybernetics}, 1997.

\bibitem{hanke2010dgs}
Thomas Hanke, Lutz K{\"o}nig, Sven Wagner, and Silke Matthes.
\newblock {DGS Corpus \& Dicta-Sign: The Hamburg Studio Setup}.
\newblock In {\em 4th Workshop on the Representation and Processing of Sign
  Languages: Corpora and Sign Language Technologies (CSLT 2010), Valletta,
  Malta}, 2010.

\bibitem{he2016deep}
Kaiming He, Xiangyu Zhang, Shaoqing Ren, and Jian Sun.
\newblock {Deep Residual Learning for Image Recognition}.
\newblock In {\em Proceedings of the IEEE Conference on Computer Vision and
  Pattern Recognition (CVPR)}, 2016.

\bibitem{huang2021towards}
Wencan Huang, Wenwen Pan, Zhou Zhao, and Qi Tian.
\newblock {Towards Fast and High-Quality Sign Language Production}.
\newblock In {\em Proceedings of the 29th ACM International Conference on
  Multimedia}, 2021.

\bibitem{jiang2021skeleton}
Songyao Jiang, Bin Sun, Lichen Wang, Yue Bai, Kunpeng Li, and Yun Fu.
\newblock {Skeleton Aware Multi-Modal Sign Language Recognition}.
\newblock In {\em Proceedings of the IEEE/CVF Conference on Computer Vision and
  Pattern Recognition (CVPR) Workshops}, 2021.

\bibitem{kao2019graph}
Jiun-Yu Kao, Antonio Ortega, Dong Tian, Hassan Mansour, and Anthony Vetro.
\newblock {Graph Based Skeleton Modeling for Human Activity Analysis}.
\newblock In {\em 2019 IEEE International Conference on Image Processing
  (ICIP)}, 2019.

\bibitem{kayahan2019hybrid}
Dilek Kayahan and Tunga G{\"u}ng{\"o}r.
\newblock {A Hybrid Translation System from Turkish Spoken Language to Turkish
  Sign Language}.
\newblock In {\em IEEE International Symposium on INnovations in Intelligent
  SysTems and Applications (INISTA)}, 2019.

\bibitem{kingma2014adam}
Diederik~P. Kingma and Jimmy Ba.
\newblock {Adam: A Method for Stochastic Optimization}.
\newblock In {\em Proceedings of the International Conference on Learning
  Representations (ICLR)}, 2014.

\bibitem{koller2015continuous}
Oscar Koller, Jens Forster, and Hermann Ney.
\newblock {Continuous Sign Language Recognition: Towards Large Vocabulary
  Statistical Recognition Systems Handling Multiple Signers}.
\newblock {\em Computer Vision and Image Understanding (CVIU)}, 2015.

\bibitem{koncel2019text}
Rik Koncel-Kedziorski, Dhanush Bekal, Yi Luan, Mirella Lapata, and Hannaneh
  Hajishirzi.
\newblock {Text Generation from Knowledge Graphs with Graph Transformers}.
\newblock {\em arXiv preprint arXiv:1904.02342}, 2019.

\bibitem{kosaraju2019social}
Vineet Kosaraju, Amir Sadeghian, Roberto Mart{\'\i}n-Mart{\'\i}n, Ian Reid,
  S~Hamid Rezatofighi, and Silvio Savarese.
\newblock {Social-BiGAT: Multimodal Trajectory Forecasting using Bicycle-GAN
  and Graph Attention Networks}.
\newblock In {\em Advances in Neural Information Processing Systems (NIPS)},
  2019.

\bibitem{kouremenos2018statistical}
Dimitris Kouremenos, Klimis~S Ntalianis, Giorgos Siolas, and Andreas
  Stafylopatis.
\newblock {Statistical Machine Translation for Greek to Greek Sign Language
  Using Parallel Corpora Produced via Rule-Based Machine Translation}.
\newblock In {\em IEEE 31st International Conference on Tools with Artificial
  Intelligence (ICTAI)}, 2018.

\bibitem{JoeyNMT}
Julia Kreutzer, Joost Bastings, and Stefan Riezler.
\newblock {Joey {NMT}: A Minimalist {NMT} Toolkit for Novices}.
\newblock In {\em Proceedings of the Conference on Empirical Methods in Natural
  Language Processing (EMNLP)}, 2019.

\bibitem{kreuzer2021rethinking}
Devin Kreuzer, Dominique Beaini, William~L Hamilton, Vincent L{\'e}tourneau,
  and Prudencio Tossou.
\newblock {Rethinking Graph Transformers with Spectral Attention}.
\newblock {\em arXiv preprint arXiv:2106.03893}, 2021.

\bibitem{li2018learning}
Yujia Li, Oriol Vinyals, Chris Dyer, Razvan Pascanu, and Peter Battaglia.
\newblock {Learning Deep Generative Models of Graphs}.
\newblock {\em arXiv preprint arXiv:1803.03324}, 2018.

\bibitem{liang1996sign}
Rung-Huei Liang and Ming Ouhyoung.
\newblock {A Sign Language Recognition System using Hidden Markov Model and
  Context Sensitive Search}.
\newblock In {\em Proceedings of the ACM Symposium on Virtual Reality Software
  and Technology}, 1996.

\bibitem{lim2019isolated}
Kian~Ming Lim, Alan Wee~Chiat Tan, Chin~Poo Lee, and Shing~Chiang Tan.
\newblock {Isolated Sign Language Recognition using Convolutional Neural
  Network Hand Modelling and Hand Energy Image}.
\newblock {\em Multimedia Tools and Applications}, 2019.

\bibitem{liu2018generating}
Peter~J Liu, Mohammad Saleh, Etienne Pot, Ben Goodrich, Ryan Sepassi, Lukasz
  Kaiser, and Noam Shazeer.
\newblock {Generating Wikipedia by Summarizing Long Sequences}.
\newblock In {\em International Conference on Learning Representations (ICLR)},
  2018.

\bibitem{lu2010collecting}
Pengfei Lu and Matt Huenerfauth.
\newblock {Collecting a Motion-Capture Corpus of American Sign Language for
  Data-Driven Generation research}.
\newblock In {\em Proceedings of the NAACL HLT 2010 Workshop on Speech and
  Language Processing for Assistive Technologies}, 2010.

\bibitem{luong2015effective}
Minh-Thang Luong, Hieu Pham, and Christopher~D Manning.
\newblock {Effective Approaches to Attention-based Neural Machine Translation}.
\newblock In {\em Proceedings of the Conference on Empirical Methods in Natural
  Language Processing (EMNLP)}, 2015.

\bibitem{mao2019learning}
Wei Mao, Miaomiao Liu, Mathieu Salzmann, and Hongdong Li.
\newblock {Learning Trajectory Dependencies for Human Motion Prediction}.
\newblock In {\em Proceedings of the IEEE/CVF International Conference on
  Computer Vision (ICCV)}, 2019.

\bibitem{meng2021attention}
Lu Meng and Ronghui Li.
\newblock {An Attention-Enhanced Multi-Scale and Dual Sign Language Recognition
  Network Based on a Graph Convolution Network}.
\newblock {\em Sensors}, 2021.

\bibitem{miyazaki2020machine}
Taro Miyazaki, Yusuke Morita, and Masanori Sano.
\newblock {Machine Translation from Spoken Language to Sign Language using
  Pre-trained Language Model as Encoder}.
\newblock In {\em Proceedings of the LREC2020 Workshop on the Representation
  and Processing of Sign Languages}, 2020.

\bibitem{paszke2017automatic}
Adam Paszke, Sam Gross, Soumith Chintala, Gregory Chanan, Edward Yang, Zachary
  DeVito, Zeming Lin, Alban Desmaison, Luca Antiga, and Adam Lerer.
\newblock {Automatic Differentiation in PyTorch}.
\newblock In {\em NIPS Autodiff Workshop}, 2017.

\bibitem{pfau2010nonmanuals}
Roland Pfau, Josep Quer, et~al.
\newblock {\em {Nonmanuals: Their Grammatical and Prosodic Roles}}.
\newblock Cambridge University Press, 2010.

\bibitem{qi2018learning}
Siyuan Qi, Wenguan Wang, Baoxiong Jia, Jianbing Shen, and Song-Chun Zhu.
\newblock {Learning Human-Object Interactions by Graph Parsing Neural
  Networks}.
\newblock In {\em Proceedings of the European Conference on Computer Vision
  (ECCV)}, 2018.

\bibitem{rodriguez2021important}
Jefferson Rodriguez and Fabio Mart{\'\i}nez.
\newblock {How Important is Motion in Sign Language Translation?}
\newblock {\em IET Computer Vision}, 2021.

\bibitem{saunders2020adversarial}
Ben Saunders, Necati~Cihan Camgoz, and Richard Bowden.
\newblock {Adversarial Training for Multi-Channel Sign Language Production}.
\newblock In {\em Proceedings of the British Machine Vision Conference (BMVC)},
  2020.

\bibitem{saunders2020everybody}
Ben Saunders, Necati~Cihan Camgoz, and Richard Bowden.
\newblock {Everybody Sign Now: Translating Spoken Language to Photo Realistic
  Sign Language Video}.
\newblock {\em arXiv preprint arXiv:2011.09846}, 2020.

\bibitem{saunders2020progressive}
Ben Saunders, Necati~Cihan Camgoz, and Richard Bowden.
\newblock {Progressive Transformers for End-to-End Sign Language Production}.
\newblock In {\em Proceedings of the European Conference on Computer Vision
  (ECCV)}, 2020.

\bibitem{saunders2021anonysign}
Ben Saunders, Necati~Cihan Camgoz, and Richard Bowden.
\newblock {AnonySign: Novel Human Appearance Synthesis for Sign Language Video
  Anonymisation}.
\newblock In {\em IEEE International Conference on Automatic Face and Gesture
  Recognition (FG) (To Appear)}, 2021.

\bibitem{saunders2021continuous}
Ben Saunders, Necati~Cihan Camgoz, and Richard Bowden.
\newblock {Continuous 3D Multi-Channel Sign Language Production via Progressive
  Transformers and Mixture Density Networks}.
\newblock In {\em International Journal of Computer Vision (IJCV)}, 2021.

\bibitem{saunders2021mixed}
Ben Saunders, Necati~Cihan Camgoz, and Richard Bowden.
\newblock {Mixed SIGNals: Sign Language Production via a Mixture of Motion
  Primitives}.
\newblock In {\em Proceedings of the International Conference on Computer
  Vision (ICCV)}, 2021.

\bibitem{shi2019skeleton}
Lei Shi, Yifan Zhang, Jian Cheng, and Hanqing Lu.
\newblock {Skeleton-based Action Recognition with Directed Graph Neural
  Networks}.
\newblock In {\em Proceedings of the IEEE Conference on Computer Vision and
  Pattern Recognition (CVPR)}, 2019.

\bibitem{song2019session}
Weiping Song, Zhiping Xiao, Yifan Wang, Laurent Charlin, Ming Zhang, and Jian
  Tang.
\newblock {Session-Based Social Recommendation via Dynamic Graph Attention
  Networks}.
\newblock In {\em Proceedings of the ACM International Conference on Web Search
  and Data Mining}, 2019.

\bibitem{stokoe1980sign}
William~C Stokoe.
\newblock {Sign Language Structure}.
\newblock {\em Annual Review of Anthropology}, 1980.

\bibitem{stoll2018sign}
Stephanie Stoll, Necati~Cihan Camgoz, Simon Hadfield, and Richard Bowden.
\newblock {Sign Language Production using Neural Machine Translation and
  Generative Adversarial Networks}.
\newblock In {\em Proceedings of the British Machine Vision Conference (BMVC)},
  2018.

\bibitem{straka2011skeletal}
Matthias Straka, Stefan Hauswiesner, Matthias R{\"u}ther, and Horst Bischof.
\newblock {Skeletal Graph Based Human Pose Estimation in Real-Time}.
\newblock In {\em Proceedings of the British Machine Vision Conference (BMVC)},
  2011.

\bibitem{sutton1999linguistics}
Rachel Sutton-Spence and Bencie Woll.
\newblock {\em {The Linguistics of British Sign Language: An Introduction}}.
\newblock Cambridge University Press, 1999.

\bibitem{tolba2012proposed}
MF Tolba, Ahmed Samir, and Magdy Abul-Ela.
\newblock {A Proposed Graph Matching Technique for Arabic Sign Language
  Continuous Sentences Recognition}.
\newblock In {\em 2012 8th International Conference on Informatics and Systems
  (INFOS)}, 2012.

\bibitem{vaswani2017attention}
Ashish Vaswani, Noam Shazeer, Niki Parmar, Jakob Uszkoreit, Llion Jones,
  Aidan~N Gomez, {\L}ukasz Kaiser, and Illia Polosukhin.
\newblock {Attention Is All You Need}.
\newblock In {\em Advances in Neural Information Processing Systems (NIPS)},
  2017.

\bibitem{velivckovic2017graph}
Petar Veli{\v{c}}kovi{\'c}, Guillem Cucurull, Arantxa Casanova, Adriana Romero,
  Pietro Lio, and Yoshua Bengio.
\newblock {Graph Attention Networks}.
\newblock In {\em Proceedings of the International Conference on Learning
  Representations (ICLR)}, 2017.

\bibitem{wilson1993neural}
Beth~J Wilson and Gretel Anspach.
\newblock {Neural Networks for Sign Language Translation}.
\newblock In {\em Applications of Artificial Neural Networks IV}. International
  Society for Optics and Photonics, 1993.

\bibitem{xiao2020skeleton}
Qinkun Xiao, Minying Qin, and Yuting Yin.
\newblock {Skeleton-based Chinese Sign Language Recognition and Generation for
  Bidirectional Communication between Deaf and Hearing People}.
\newblock In {\em Neural Networks}, 2020.

\bibitem{yan2018spatial}
Sijie Yan, Yuanjun Xiong, and Dahua Lin.
\newblock {Spatial Temporal Graph Convolutional Networks for Skeleton-Based
  Action Recognition}.
\newblock In {\em Proceedings of the AAAI Conference on Artificial
  Intelligence}, 2018.

\bibitem{yang2018modeling}
Baosong Yang, Zhaopeng Tu, Derek~F Wong, Fandong Meng, Lidia~S Chao, and Tong
  Zhang.
\newblock {Modeling Localness for Self-Attention Networks}.
\newblock In {\em Proceedings of the 2018 Conference on Empirical Methods in
  Natural Language Processing (EMNLP)}, 2018.

\bibitem{yang2019conditional}
Carl Yang, Peiye Zhuang, Wenhan Shi, Alan Luu, and Pan Li.
\newblock {Conditional Structure Generation through Graph Variational
  Generative Adversarial Nets}.
\newblock In {\em Advances in Neural Information Processing Systems (NIPS)},
  2019.

\bibitem{yao2019graph}
Liang Yao, Chengsheng Mao, and Yuan Luo.
\newblock {Graph Convolutional Networks for Text Classification}.
\newblock In {\em Proceedings of the AAAI Conference on Artificial
  Intelligence}, 2019.

\bibitem{yin2020sign}
Kayo Yin.
\newblock {Sign Language Translation with Transformers}.
\newblock {\em arXiv preprint arXiv:2004.00588}, 2020.

\bibitem{yu2020spatio}
Cunjun Yu, Xiao Ma, Jiawei Ren, Haiyu Zhao, and Shuai Yi.
\newblock {Spatio-Temporal Graph Transformer Networks for Pedestrian Trajectory
  Prediction}.
\newblock In {\em Proceedings of the European Conference on Computer Vision
  (ECCV)}, 2020.

\bibitem{yun2019graph}
Seongjun Yun, Minbyul Jeong, Raehyun Kim, Jaewoo Kang, and Hyunwoo~J Kim.
\newblock {Graph Transformer Networks}.
\newblock In {\em Advances in Neural Information Processing Systems (NIPS)},
  2019.

\bibitem{zelinka2020neural}
Jan Zelinka and Jakub Kanis.
\newblock {Neural Sign Language Synthesis: Words Are Our Glosses}.
\newblock In {\em The IEEE Winter Conference on Applications of Computer Vision
  (WACV)}, 2020.

\bibitem{zhang2016chinese}
Jihai Zhang, Wengang Zhou, Chao Xie, Junfu Pu, and Houqiang Li.
\newblock {Chinese Sign Language Recognition with Adaptive HMM}.
\newblock In {\em 2016 IEEE International Conference on Multimedia and Expo
  (ICME)}, 2016.

\bibitem{zhao2019explicit}
Guangxiang Zhao, Junyang Lin, Zhiyuan Zhang, Xuancheng Ren, Qi Su, and Xu Sun.
\newblock {Explicit Sparse Transformer: Concentrated Attention Through Explicit
  Selection}.
\newblock {\em arXiv preprint arXiv:1912.11637}, 2019.

\bibitem{zhou2020graph}
Jie Zhou, Ganqu Cui, Shengding Hu, Zhengyan Zhang, Cheng Yang, Zhiyuan Liu,
  Lifeng Wang, Changcheng Li, and Maosong Sun.
\newblock {Graph Neural Networks: A Review of Methods and Applications}.
\newblock {\em AI Open}, 2020.

\end{thebibliography}
}

\end{document}